%% file: root.tex
\documentclass[letterpaper, 10 pt, conference]{ieeeconf}  

\IEEEoverridecommandlockouts                              

\input{macros}
\title{\LARGE \bf
PeRP: Personalized Residual Policies For Congestion Mitigation \\Through Co-operative Advisory Systems
}

\author{Aamir Hasan$^{1}$, Neeloy Chakraborty$^{*1}$, Haonan Chen$^{*1}$, Jung-Hoon Cho$^{2}$, \\Cathy Wu$^{2}$, and Katherine Driggs-Campbell$^{1}$
\thanks{* denotes equal contribution.}
\thanks{$^{1}$A. Hasan, N. Chakraborty, H. Chen, and K. Driggs-Campbell are with the Department of Electrical and Computer Engineering at the University of Illinois Urbana-Champaign. Emails: {\tt\small \{aamirh2, neeloyc2, haonan2, krdc\}@illnois.edu}}%
\thanks{$^{2}$J.-H. Cho is with the MIT Laboratory for Information \& Decision Systems (LIDS) and the Department of Civil and Environmental Engineering (CEE) at the Massachusetts Institute of Technology, Email: {\tt\small jhooncho@mit.edu}}%
\thanks{$^{3}$C. Wu is with the MIT Laboratory for Information \& Decision Systems (LIDS), the Department of Civil and Environmental Engineering (CEE), and the Institute for Data, Systems, \& Society (IDSS) at the Massachusetts Institute of Technology, Email: {\tt\small cathywu@mit.edu}}%
}

\begin{document}

\maketitle
\thispagestyle{empty}
\pagestyle{empty}

\input{sections/00_abstract}

\input{sections/01_introduction}
\input{sections/02_related_work}
\input{sections/03_method}
\input{sections/04_experiments}
\input{sections/05_discussion}
\input{sections/06_conclusion}





\input{sections/09_appendix}


\bibliographystyle{IEEEtran}
\bibliography{root}

\addtolength{\textheight}{-14cm}   

\end{document}


\maketitle
\thispagestyle{empty}
\pagestyle{empty}


\begin{table}[!t]
    \centering
    \caption{Results comparing the performance of our baselines with PeRP}
    \begin{tabular}{l  c c }
        \toprule
         Policy & 
         \multicolumn{2}{c}{$\delta=1~(0.1\text{s})$}  \\ 
         & Avg. Speed$(\uparrow)$& Avg. Std$(\downarrow)$\\
         \midrule
         Optimal & 
         8.650 \\
         OSL & 
         6.764 & 1.768 \\
         PCP & 
         8.033 & \textbf{1.596} \\ 
         V-RP &
         8.054 & 1.667 \\ 
         TA-RP   & 
         8.153 & 1.612 \\ 
         \textbf{PeRP$^*$} & 
         \textbf{8.169} & 1.617 \\ 
        \bottomrule
    \end{tabular}
    \label{tab:quantitative_1}
\end{table}

\begin{table*}[!b]
    \centering
    \caption{Comparing the performance of our baselines with PeRP for the total emissions released during evaluation. All values are to the order of $10^8$}
    \begin{tabular}{l  c c c c c }
        \toprule
         Policy & 
         $\delta=1~(0.1\text{s})$ &
         $\delta=10~(1\text{s})$ & 
         $\delta=20~(2\text{s})$ & 
         $\delta=50~(5\text{s})$ &
         $\delta=100~(10\text{s})$ \\
         \midrule
         PCP & 
         3.028 $\pm$ 0.319 &  
         \textbf{3.006 $\pm$ 0.044} & 
         \textbf{3.036 $\pm$ 0.085} & 
         3.015 $\pm$ 0.103 &
         3.065 $\pm$ 0.459 \\
         V-RP & 
         \textbf{3.020 $\pm$ 0.317} &  
         3.064 $\pm$ 0.082 & 
         3.044 $\pm$ 0.081 & 
         2.945 $\pm$ 0.064 &
         2.955 $\pm$ 0.433 \\
         TA-RP & 
         3.061 $\pm$ 0.097 &  
         3.083 $\pm$ 0.074 & 
         3.044 $\pm$ 0.077 & 
         2.960 $\pm$ 0.065 &
         2.916 $\pm$ 0.523 \\
         \textbf{PeRP$^*$} & 
         3.048 $\pm$ 0.081 &  
         3.064 $\pm$ 0.079 & 
         3.062 $\pm$ 0.083 & 
         \textbf{2.938 $\pm$ 0.301} &
         \textbf{2.913 $\pm$ 0.521} \\
        \bottomrule
    \end{tabular}
    \label{tab:quantitative_emissions}
\end{table*}

\begin{table*}[!b]
\centering
    \caption{Frequency of collisions during evaluations for baselines with PeRP}
    \begin{tabular}{l  c c c c c}
        \toprule
         Policy & 
         $\delta=1~(0.1\text{s})$ &
         $\delta=10~(1\text{s})$ & 
         $\delta=20~(2\text{s})$ & 
         $\delta=50~(5\text{s})$ &
         $\delta=100~(10\text{s})$ \\
         \midrule
         PCP & 
         2 &  
         4 & 
         \textbf{0} & 
         \textbf{0} &
         \textbf{4} \\
         V-RP & 
         2 &  
         \textbf{0} & 
         \textbf{0} & 
         \textbf{0} &
         \textbf{4} \\
         TA-RP & 
         \textbf{0} &  
         \textbf{0} & 
         \textbf{0} & 
         \textbf{0} &
         6 \\
         \textbf{PeRP$^*$} & 
         \textbf{0} &  
         \textbf{0} & 
         \textbf{0} & 
         2 &
         5 \\
        \bottomrule
    \end{tabular}
    \label{tab:quantitative_collisions}
\end{table*}

\section{Additional Results}
Table \ref{tab:quantitative_1} shows the evaluation metrics for the hold length $\delta=1 (0.1s)$.
As with the other hold-lengths, PeRP achieves the best overall mean speed but is more volatile in its performance.
However, these results are not particularly important due to the impracticality of using a hold-length of $0.1$s. 

Table \ref{tab:quantitative_emissions} compares the policies on the total carbon emissions released during evaluation trials. 
The total carbon emissions released in the cases for hold-lengths $\delta={1, 10, 20}$ are comparable for all policies only differing 3rd decimal place. 
However, significant differences are noticeable in the cases for larger hold lengths ($\delta={50, 100}$) where PeRP outperforms the baseline models.

Table \ref{tab:quantitative_collisions} provides the frequency of collisions for the evaluation trials. 
Most policies learn to avoid collisions across all hold lengths, with all residual policies improving on the performance of the PC policies. 
Note that the frequency of collision decreases around the optimal hold length and then increases for larger hold lengths.

\addtolength{\textheight}{-12cm}   

%% file: macros.tex
\usepackage{amsmath}
\usepackage{amssymb}
\usepackage{stmaryrd}
\usepackage{mathtools}
\usepackage{acronym}
\usepackage{color}
\usepackage{xcolor}
\usepackage{verbatim}
\usepackage{booktabs}
\usepackage{siunitx}
\usepackage{graphics}
\usepackage{graphicx,caption}
\usepackage{suffix}
\usepackage{xstring}
\usepackage{xparse}
\usepackage{expl3}
\usepackage{mathrsfs}
\usepackage{tabularx}
\usepackage{makecell}
\usepackage{array}
\usepackage{hyperref}
\usepackage{cleveref}

\usepackage[utf8]{inputenc} 
\usepackage[T1]{fontenc}    
\usepackage{amsfonts}       
\usepackage{microtype}      
\usepackage{cite}
\AtBeginEnvironment{quote}{\singlespacing\small}
\usepackage{dirtytalk}
\usepackage{lscape}

\usepackage[ruled]{algorithm2e}

\newcommand{\etal}{\textit{et al}.}
\newcommand{\ie}{\textit{i}.\textit{e}.}
\newcommand{\eg}{\textit{e}.\textit{g}.}

\usepackage{subcaption}
\usepackage{wrapfig}

\usepackage{tikz}
\usetikzlibrary{arrows,backgrounds,calc}
\usepackage{relsize}
\usepackage{float}
\usepackage{kantlipsum} 
\usepackage{lipsum}
\usepackage{stfloats}

\usepackage{todonotes}
\usepackage{soul}
\definecolor{smoothgreen}{rgb}{0.7,1,0.7}
\sethlcolor{smoothgreen}

\RequirePackage{luatex85}
\usepackage{pgfplots}
\pgfplotsset{compat=newest}
\pgfplotsset{every axis legend/.append style={%
		cells={anchor=west}}
}
\usepgfplotslibrary{polar}
\usetikzlibrary{arrows}
\tikzset{>=stealth'}

\definecolor{C1}{rgb}{0.0, 0.447, 0.741}
\definecolor{C1_light}{rgb}{0.0, 0.6032388663967612, 1.0}
\definecolor{C2}{rgb}{0.85, 0.325, 0.098}
\definecolor{C3}{rgb}{0.929, 0.694, 0.125}
\definecolor{C4}{rgb}{0.494, 0.184, 0.556}
\definecolor{C5}{rgb}{0.466, 0.674, 0.188}
\definecolor{C6}{rgb}{0.301, 0.745, 0.933}
\definecolor{C7}{rgb}{0.635, 0.078, 0.184}

\usepgfplotslibrary{groupplots}

\usetikzlibrary{shapes.geometric, arrows}

\tikzstyle{startstop} = [rectangle, rounded corners, minimum width=2cm, minimum height=1cm,text centered, draw=black, fill=none]
\tikzstyle{arrow} = [thick,->,>=stealth]

%% file: sections/00_abstract.tex
\begin{abstract}
Intelligent driving systems can be used to mitigate congestion through simple actions, thus improving many socioeconomic factors such as commute time and gas costs. 
However, these systems assume precise control over autonomous vehicle fleets, and are hence limited in practice as they fail to account for uncertainty in human behavior. 
Piecewise Constant (PC) Policies address these issues by structurally modeling the likeness of human driving to reduce traffic congestion in dense scenarios to provide action advice to be followed by human drivers.
However, PC policies assume that all drivers behave similarly.
To this end, we develop a co-operative advisory system based on PC policies with a novel driver trait conditioned Personalized Residual Policy, PeRP. 
PeRP advises drivers to behave in ways that mitigate traffic congestion. 
We first infer the driver’s intrinsic traits on how they follow instructions in an unsupervised manner with a variational autoencoder. 
Then, a policy conditioned on the inferred trait adapts the action of the PC policy to provide the driver with a personalized recommendation.
Our system is trained in simulation with novel driver modeling of instruction adherence.
We show that our approach successfully mitigates congestion while adapting to different driver behaviors, with 4 to 22\% improvement in average speed over baselines. 
\footnote{Additional material and code is available at the project webpage: \url{https://sites.google.com/illinois.edu/perp}}
\end{abstract}

%% file: sections/01_introduction.tex
\section{Introduction}
\label{sec:intro}


Traffic congestion is a leading cause of urban mobility issues such as long commute times and increased fuel consumption~\cite{epa, emissions}.
These problems also have a severe economic impact on everyday living~\cite{goodwin2004economic}.
Thus, alleviating congestion (\ie~congestion mitigation) is paramount to improving multiple facets of modern society.
Simple speed management techniques through speed and acceleration advice have been shown to improve safety while reducing congestion~\cite{french, wilmot1999}.
However, these methods utilize generic speed limits that are independent of important dynamic factors such as the number of vehicles on the road.

Autonomous vehicles~(AVs) trained to operate optimally can also be used for congestion mitigation~\cite{flow}.
For example, Stern \etal~showed that a single AV can reduce stop-and-go waves and thus reduce emissions in a closed-loop traffic setting~\cite{stern2018}.
However, current AV systems are immature in non-idealized settings, particularly in scenarios involving a large number of agents that are ubiquitous in traffic scenarios~\cite{Kalra2014, dixit2016autonomous}.
The same policies that control AVs can be modified for use in shared control schemes to aid in congestion mitigation~\cite{shared_control_survey}.
Particularly, co-operative advisory systems where drivers are provided instructions by these optimally trained policies can be utilized to provide near optimal performance~\cite{pcp}.
These policies can utilize simple methods to interact with drivers such as direct integration into smartphone applications (\eg~navigation apps) or through in-car interfaces (\eg~head-up displays).

\begin{figure}[tb!]
    \centering
    \includegraphics[width=\columnwidth]{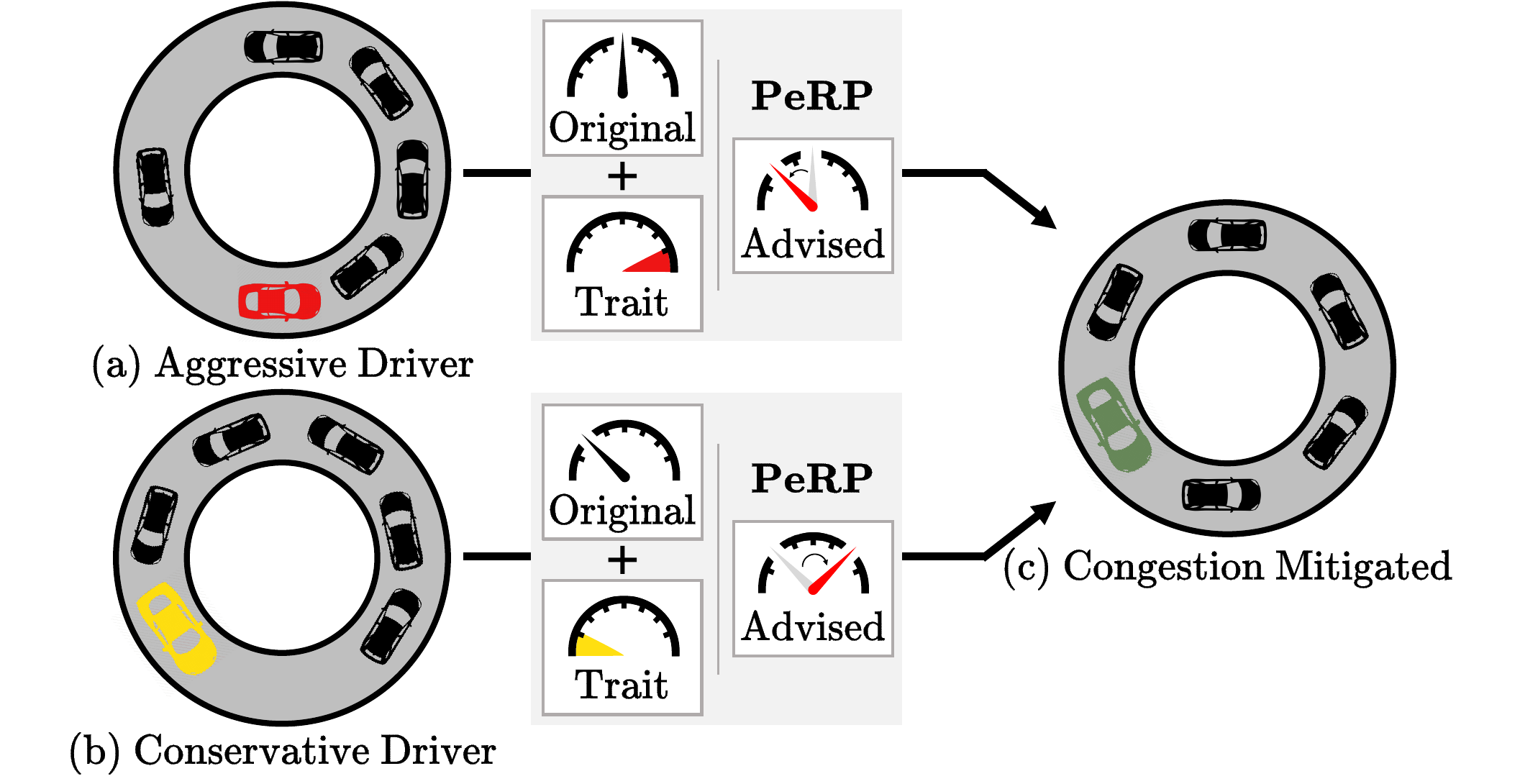}
    \caption{PeRP augments general advised instructions to provide personalized recommendations to drivers of varying driving styles (\eg~an aggressive driver \textbf{(a)} or conservative driver \textbf{(b)}) to mitigate congestion \textbf{(c)}.}
    \label{fig:eye-catching}
\end{figure}

AV policies trained with Reinforcement Learning (RL) can also successfully stabilize traffic flow and mitigate the formation of stop-and-go waves~\cite{flow, emergent}.
For example, Sridhar \etal~ use piecewise constant (PC) policies to mitigate congestion by explicitly learning \emph{human compatible} policies~\cite{pcp}.
This modeling is achieved through their PC constraint, where actions are held for some duration, defined as the hold-length, to allow drivers to adjust their action to the provided advice.
While these policies are designed to be easy to follow, they assume a one-solution-for-all paradigm where all drivers behave similarly.
However, we can adapt these PC policies to produce personalized residual policies that account for variances in human behavior. 


To this end, we propose a class of \textbf{Pe}rsonalized \textbf{R}esidual \textbf{P}olicies, PeRP, that condition the advised action on an inferred driver trait.
Figure \ref{fig:eye-catching} shows an overview of our PeRP.
We utilize a variational autoencoder to perform unsupervised trait inference on how drivers follow instructions.
The inferred driver traits and actions by the base policy are used to create a personalized advised action that can be followed by drivers without significant load to mitigate congestion.
Thus, making our policy agnostic to different driver behaviors.
We verify the effectiveness of our policy in simulation using a novel driver instruction-following model.
Most notably, we model how drivers follow advice imperfectly.

Our main contributions are:
\begin{enumerate}
    \item A novel Personalized Residual Policy for congestion mitigation through co-operative advisory systems.
    \item A simulation framework for modeling driver responses to advisory systems in on-road settings\footnotemark[1].
    \item A novel unsupervised driver trait inference model that encodes how drivers follow advice provided to them.
\end{enumerate}

This paper is organized as follows. 
In Section \ref{sec:related_work}, we describe relevant work. 
Section \ref{sec:prelim} contains preliminary information required to aid in understanding our proposed method that is detailed in Section \ref{sec:method}. 
We present our experiments and discuss results in Sections~\ref{sec:experiments} and~\ref{sec:results}, respectively. 
Finally, we provide our conclusions and avenues for future work in Section \ref{sec:conclusion}.  

%% file: sections/02_related_work.tex
\section{Related Work}
\label{sec:related_work}


\subsection{RL for designing Congestion Mitigation Strategies}
With the advancement of model-free deep RL, many policies have been developed to mitigate traffic congestion and reduce emissions. 
These policies are used to improve throughput in traffic systems~\cite{kreidieh_dissipating_2018, vinitsky_lagrangian_2018, yan_reinforcement_2021, flow} and developing eco-driving strategies in a myriad of road settings~\cite{ jayawardana2022ecolearning, jayawardana_mixed_2021, wegener_automated_2021}. 
Particularly, Jayawardana \etal~design interpretable decision tree policies and supervision models that perform at the same level as the RL based policies to mitigate congestion in urban scenarios~\cite{jayawardana_mixed_2021}.
Alternatively, Wu \etal~propose a general modular learning framework to analyze the impact of AVs on traffic flow~\cite{flow}.
To the same effect, Sridhar and Wu introduce PC policies to mitigate congestion and stabilize traffic~\cite{pcp}.
PC policies achieve human compatibility by using the notion of an action extension parameter, where an action is held for some guidance hold-length, $\delta$, that dictates the control frequency~\cite{metelli_control_2020}. 
Theoretical analysis on such human-compatible guidance derives conditions that directly relate the advice provided to the drivers with the stability of traffic flow~\cite{li_integrated_2023}.
While these PC policies are immensely useful, they assume that all drivers co-operate and react to the guidance similarly.
In this work, we address these caveats by deriving residual policies that use learned driver reactions to the given guidance.
We provide a summary of PC policies in Section \ref{sec:method-pcp} as they are the base policies for our class of residual policies.

\subsection{Residual Policy Learning}

Residual Policy Learning (RPL) is grounded in two main facets: 
(1) utilising a base policy to produce an approximate solution 
 and 
(2) learning a corrective term to mitigate inaccuracies in the base policy and handle potential variations~\cite{10.1145/1143844.1143845, pmlr-v168-jiang22a, 10.1109/TRO.2020.2988642}.
RPL represents a biased exploration strategy towards the state distribution of the initial policy as it is used to learn corrective values for control parameters or actions~\cite{Ajay2018AugmentingPS, schaff2020, silver2018, johannik2019}.
For example, Rana \etal~employ RPL for low-level controllers to achieve fine-grained skill adaptation, thus enabling downstream RL agents to adapt to unseen environment variations~\cite{pmlr-v205-rana23a}. 
Similarly, Zhang \etal~utilize RPL to enhance a modified artificial potential field (MAPF) policy for high-speed autonomous racing that uses RPL to guide exploration~\cite{9834085}. 
However, the residual policies are not adaptable as they do not assimilate to controllers that have minor differences from the trained controller. 
In contrast, our residual policy, PeRP, is conditioned on a learned driver trait.
This conditioning enables the adjustment of the output action based on different driver behaviors and can thus effectively handle the nuances in human-autonomy teaming. 

\subsection{Driver Trait Inference}

Factors such as driving experience, personal preferences, fatigue, and levels of distraction impact how people drive~\cite{brown2020taxonomy}.
As it is paramount to tailor driver-centric autonomous systems to these factors, Driver Trait Inference (DTI) is a booming field of study with applications in driver intent estimation~\cite{Deo2018intent, Gillmeier2019intent}, vehicle trajectory prediction~\cite{huang2020long, Liu2019motion}, and mulit-agent path planning~\cite{morton2017, liu2021learning, ma2020reinforcement}. 
%
Popular approaches to DTI can be decomposed into combinations of supervised and unsupervised methods.
Supervised approaches 
require discrete labels during training which significantly hinders the accuracy of the model during evaluation, as drivers may act with out of distribution traits at test time~\cite{ma2020reinforcement}. 
In contrast, unsupervised learning methods fit a continuous distribution over unlabeled driver traits from trajectories and show that a compressed latent representation can inform lower-level planning~\cite{morton2017, liu2021learning}.
Other works use a combination of unsupervised learning and RL to estimate the strategies or traits of other agents for better interactions~\cite{lili, wang2021influencing, losey_rili}.
We chose to preform DTI with an Variational Autoencoder based model as unsupervised methods can handle out of distribution inputs without significant errors while being more data efficient~\cite{vae, liu2021learning}.

In early experiments, we found that the latent encodings on the inference of context-aware behavioral traits was more useful for downstream RL policies than the inference of Intelligent Driver Model~(IDM)~\cite{idm} parameters that are used in works by Morton \etal~\cite{morton2017}. 
Thus, our DTI models the preferred action offset from the advised action output by the base policy.
Intuitively, when drivers are provided speed advice, they will either be (1) conservative and drive slower than the advised speed, (2) aggressive and drive over the advised speed, or (3) follow the advised speed as closely as possible.
Our trait inference model is thus designed to capture this phenomena to inform the downstream personalizaton task. 


%% file: sections/03_method.tex
\section{Preliminaries}
\label{sec:prelim}

\subsection{Problem Definition}
\label{sec:method-prob-def}
Consider a road network with $N$ vehicles, where a single ego vehicle is driven by a human following some guidance, while $N-1$ vehicles are driven by humans without any advice.
Given the current state of the ego vehicle $s \in S$ and an advised action $a^{advised} \in A$, the driver of the ego vehicle takes action $a^{driver} \in A$ according to some policy $\pi^{driver}:~S~\times A~\rightarrow~A$, where $S$ and $A$ are the predefined state and action spaces for the road network, respectively.
Similarly, each driver $i$ of the $N-1$ vehicles, takes in the current state of their vehicle $s^{i} \in S$ and applies an action $a^{i} \in A$ according to a policy $\pi^{i}(s): S \rightarrow A$.
$$a^{driver} \sim \pi^{driver}(s, a^{advised})$$
$$ a^{i} \sim \pi^{i}(s) \forall i \in \{1, ..., N-1\}$$
We seek to find a policy, $\pi(s): S \rightarrow A$ that provides the advised action, $a^{advised}$, to the ego driver that mitigates congestion in the network. 
In this work, we choose to model the action as a speed action.
Without loss of generality, the same techniques can be applied to acceleration actions as the system is agnostic to the type of action.

Theoretically, policies that are dependent on the network parameters that produce a constant speed action can mitigate congestion~\cite{stern2018, flow}.
However, these policies assume static environments and are not practically permissible to be followed by drivers due to the nuances of human behavior (\eg~reaction time).
We encapsulate these diverse behaviors by the driver policy, $\pi^{driver}$.
Note that for the remainder of this paper, we refer to the ego vehicle and its driver as the driver or agent interchangeably.

\subsection{Piecewise Constant Policies (PCP)}
\label{sec:method-pcp}


Humans require a few seconds ($\approx 2-3s$) to perceive and act on the instructions provided to them~\cite{Droździel2020, Jurecki2017}.
PC policies aim to incorporate this delay in action propagation by holding an action for a specified 
length of time, $\delta$.
Formally, PC policies for providing instructions to drivers can be defined as an episodic Markov Decision Process (MDP).
An episodic MDP: $\mathcal{M}$, is defined as, $\mathcal{M}=(S, A, P, R, H, \delta, \gamma)$, where $S$ is the state space, $A$ is the action space, $P: S \times A \rightarrow S$ represents the transition probabilities, $r(a, s) \in R: S \times A \rightarrow \mathbb{R}$ is the reward function for a state-action pair $(s, a)$, $H$ represents the horizon, $\delta$ represents the number of timesteps an action is held, and $\gamma \in [0, 1)$ is the discount factor.

For PC policies we define the state $s \in S^{PCP}$, with regards to the ego vehicle as: 
$$s=\left(\frac{v_{ego}}{v_{max}}, \frac{v_{leader}}{v_{max}}, \frac{h_{leader}}{h_{max}}\right)$$
where $v_{ego}$ is the speed of the ego vehicle, $v_{leader}$ is the speed of the vehicle leading the ego vehicle, $h_{leader}$ is the headway distance between the ego vehicle and its leader, $v_{max}$ is the maximum speed allowed on the track, and $h_{max}$ is the maximum possible distance between the two vehicles. 
We assume a fully observable state, $s \in S^{PCP}$, for the agent.
The action space for PC policies is defined as $A^{PCP}=\{0, ..., A_{max}\}$, a discrete set of $\alpha$ equally spaced speeds in m/s, where $\alpha \text{ and } A_{max}$ are hyperparameters. 
The reward function used for the PC policies is the speed of the ego vehicle, $R^{PCP} = v_{ego} = a^{driver}$.
We assume the driver has exact control over the speed of the ego vehicle.
The average speed of all vehicles would also function as a good reward function as the two definitions would produce similar policies. 
However, the latter definition assumes full observability of the environment and is thus unrealistic. 

At each timestep, $t$, the PC policy, $\pi^{PCP}(s_t): S^{PCP} \rightarrow A^{PCP}$, chooses an action $a^{PCP}_t \in A^{PCP}$ that represents the speed to be held for $\delta$ timesteps that maximizes the reward function, $R^{PCP}$.
We refer the readers to the work by Sridhar \etal~and Li \etal~for an in-depth analysis of PC policies and their robust nature~\cite{pcp, li_integrated_2023}.

While the above definition uses speed as its action space, the original work by Sridhar \etal~uses acceleration. 
We choose speed as our action space since humans find it easier to perceive and follow advice in terms of speed over acceleration~\cite{hcd-workshop}. 
This change in action type retains all theoretical claims made about the robustness of PC Policies.


Additionally, PC policies assume that human drivers follow instructions presented to them perfectly, \ie~
$\pi^{driver}(s, a^{PCP}) = a^{PCP}$.
While this assumption alleviates training and modeling concerns, it is not pragmatic.
Studies conducted on the perception and effectiveness of speed limits have shown that drivers do not follow advice perfectly~\cite{mannering2009}, but instead offset from the advised speeds depending on the nature of the driving scenario.
We posit that these different driving styles should be considered while designing advisory systems.
Thus, we introduce a driver trait inference module to capture the variations in how drivers follow instructions. 

\section{Method}
\label{sec:method}

\begin{figure}[t!]
    \centering
    \includegraphics[width=\columnwidth]{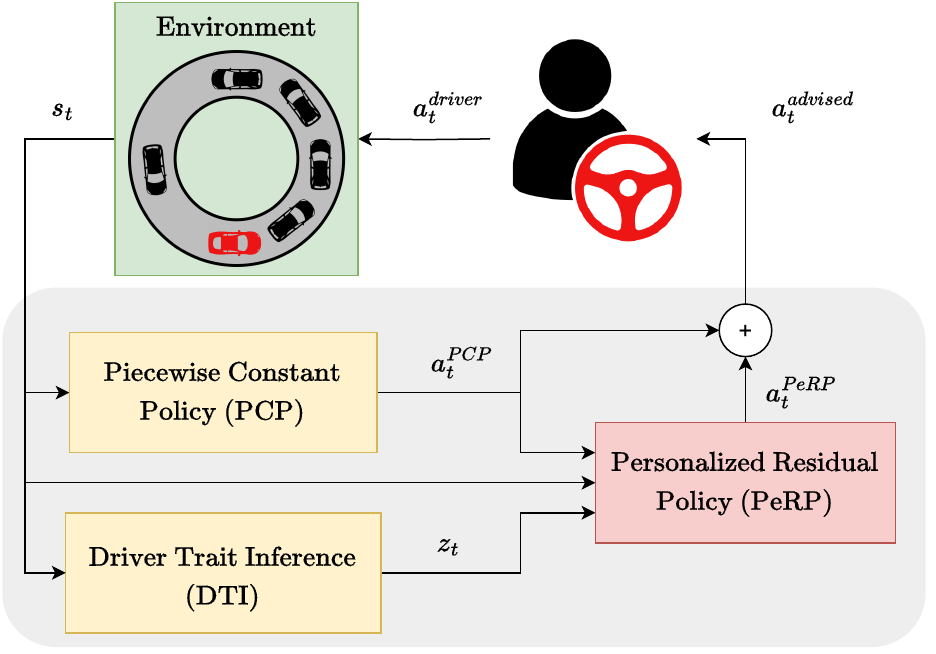}
    \caption{An overview of our co-operative advisory system: PeRP appends a residual action, $a^{PeRP}$, to the PCP action, $a^{PCP}$, while conditioned on the driver trait, $z$, to produce an advised action, $a^{advised}$. The driver considers the advised action and takes an action $a^{driver}$ in the environment.}
    \label{fig:block-diagram}
\end{figure}

In this section we describe the different modules in our proposed system.
First, we introduce driver traits and summarize our trait inference model.
Then, we introduce PeRP, our personalized residual policy for co-operative advisory systems.
Figure \ref{fig:block-diagram} shows the general architecture of our proposed system.

\subsection{Driver Policy Model}
\label{sec:method-driver-policy}

The inference of driving styles benefits advisory systems as all drivers do not behave identically~\cite{brown2020taxonomy, idm, morton2017, liu2021learning}. 
A survey conducted by Mannering confirmed that drivers do not strictly adhere to speed limits, but instead drive at higher speeds~\cite{mannering2009}.
Motivated by this case of speed limits, we define our driver trait as the drivers' preferred offset from the advised speed action.
In our case, we capture whether the driving speed is 2.5 or 5 m/s ($\approx$5 or 10 mph) over or under the advised action.
Specifically, the traits are: if the driver is driving \{5m/s below, 2.5m/s below, at, 2.5m/s above, 5m/s above\} the advised speed.

We choose these five different traits to encompass the instruction following behaviors of a majority of drivers. 
As drivers cannot maintain perfect speeds, we model these traits as deviations from the advised action with Gaussian distributions with the means $\text{trait}_\mu = \{-5, -2.5, 0, 2.5, 5\}$ and variance 1, respectively. 
Formally, this can be defined as:
$$\pi^{driver}(s, a^{advised}) = a^{advised} + k \text{ where } k \sim \mathcal{N}(\text{trait}_\mu, 1)$$

We assume perfect driver reaction time but imperfect instruction following. 
We abstain from including driver reaction times in the traits in order to limit the scope of the personalization.
Additionally, we assume that all other drivers on the road behave similarly and follow an Intelligent Driver Model (IDM)~\cite{idm}.
We believe this is a justified assumption as the main focus of this work is to mitigate congestion using a single driver and their trait, while excluding all other drivers in the system.
However, our model can easily be modified to assign different traits for all drivers. 

\subsection{Driver Trait Inference (DTI)}
\label{sec:method-trait-inference}

\begin{figure}[tb!]
    \centering
    \includegraphics[width=\columnwidth]{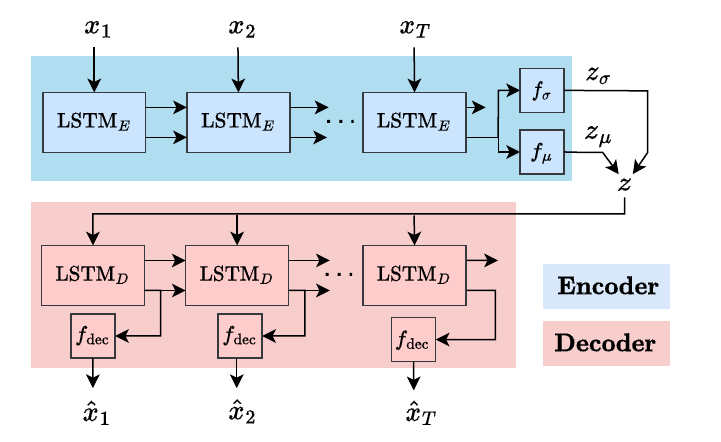}
    \caption{The Driver Trait Inference VAE Model. The input trajectory $\boldsymbol{x}$ is encoded as $z_\mu$ and $z_\sigma$ before parameterization as $z$ using the encoder network. The decoder network uses this latent vector $z$ to reconstruct the input trajectory as $\boldsymbol{\hat{x}}$.}
    \label{fig:vae_trait_inference}
\end{figure}

We capture the different driver traits discussed above in an unsupervised manner through a Variational Autonecoder (VAE) based model~\cite{vae}.
Figure \ref{fig:vae_trait_inference} summarizes the architecture of our VAE.
The input to the DTI model is a sequence of states of the agent, $\boldsymbol{x}=\{x_1, x_2, .., x_T\}$, where each $x_t \in S^{PCP}$ and $T$ is the observation period.
The VAE model first uses an encoder, consisting of a recurrent $\text{LSTM}_E$ layer and two fully connected layers, $f_\mu$ and $f_\sigma$, to encode the input trajectory $\boldsymbol{x}$ to a mean vector, $z_\mu$, and a log variance vector, $z_\sigma$.
$$z_\mu = f_\mu(e) \text{ and } z_\sigma = f_\sigma(e), \text{ where } e = \text{LSTM}_E(\boldsymbol{x})$$
We parameterize the latent vector using $z_\mu$ and $z_\sigma$ to produce $z \in \mathbb{R}^{l}$, where $l$ is the latent space dimension~\cite{vae}.
$$z = z_\mu + \epsilon \cdot exp(z_\sigma), \text{ where } \epsilon \sim \mathcal{N}(0, 1)$$
During training, the latent vector is processed with a decoder, consisting of a recurrent $\text{LSTM}_D$ layer and a fully connected layer, $f_{dec}$, to produce a reconstructed trajectory $\boldsymbol{\hat{x}} = \{\hat{x}_1, \hat{x}_2, ... , \hat{x}_T\}$.
Each element $\hat{x}_t$ is given by:
$$\hat{x}_t = f_{dec}(h_t)$$
where $h_t$ is the hidden state from $\text{LSTM}_D$ at time $t$.
We choose LSTMs for the recurrent layers in the model due to their capability learning patterns in sequences while handling variable-length sequences~\cite{hasan2022metapath, huang2020long}.

The VAE model is encouraged to reconstruct the input trajectory while ensuring that the latent vectors follow a standard normal distribution by optimising 
the loss function:
$$\mathcal{L}_{DTI} = \beta_{\text{recon}} \cdot {\left\| \boldsymbol{\hat{x}} - \boldsymbol{x} \right\|}_2 + \beta_{\text{KL}} \cdot D_{\text{KL}}(z_\mu,z_\sigma)$$ where $D_{\text{KL}}(\mu, \sigma)$ is the KL divergence between any Gaussian distribution $\mathcal{N}(\mu, \sigma)$ and the standard normal distribution $\mathcal{N}(0,1)$, and $\beta_{\text{recon}}$ and $\beta_{\text{KL}}$ are hyperparameters.

During evaluation and usage with PeRP, the decoder network is omitted.
Therefore, at each timestep $t$, the DTI module, $f_{trait}: \mathbb{R}^{T\times|S^{PCP}|} \rightarrow \mathbb{R}^l$ uses the sequence of states, $\boldsymbol{x_t}=\{x_{t-T}, x_{t^- (T-1)}, .., x_{t}\}$, to produce a latent vector $z_t$ that indicates the driver's trait.

\subsection{Personalized Residual Policies (PeRP)}

The inference of the driver's trait allows us to learn a residual policy from PC policies to provide personalized advice.
The same definition of an episodic MDP, $\mathcal{M}$, from Section \ref{sec:method-pcp} can be used to define our Personalized Residual Policies with the following modifications.
The state, $s^{PeRP} \in S^{PeRP} = S^{PCP} \times A^{PCP} \times \mathcal{L}$, is a (3 + 1 + $l$) dimensional tuple with observed state $s \in S^{PCP}$, the PCP action, $a^{PCP} \in A^{PCP}$, and the driver trait, $z \in \mathbb{R}^l$.
$$s^{PeRP}= \left(s, a^{PCP}, z\right)$$
The action space is a continuous bounded real space $A^{PeRP} = [-\epsilon, \epsilon]$, where $\epsilon$ is a hyperparameter.
The reward function, $R^{PeRP}$ is defined as:
$$R^{PeRP} = a^{driver} - |a^{driver} - a^{PCP}|$$
The first term in this reward function aims at ensuring the dissipation of stop-and-go traffic waves, similar to $R^{PCP}$.
The second term in the reward function aims at reducing the difference between the action performed by the driver and the PCP action.
We posit that an action that reduces this difference, ensures independence to driving styles and thus allows for a more co-operative advisory system.


At each timestep, $t$, the PeRP, $\pi^{PeRP}(s_t, a^{PCP}_t, z_t): S^{PeRP} \rightarrow A^{PeRP}$, chooses a residual action $a^{PeRP}_t \in A^{PeRP}$.
Then the action to be held for $\delta$ timesteps to mitigate congestion is $a^{advised}_t = a^{PCP}_t + a^{PeRP}_t$.
This advised action, $a^{advised}_t$, can be provided to the driver to reduce congestion and thereby improve emission rates in a co-operative manner.
See Algorithm \ref{alg:perp} for more details.

To train and evaluate both PC policies and PeRP, we employ a warm-up period of $W$ timesteps in each rollout. 
During the warm-up period, all vehicles are controlled by IDM to generate stop-and-go traffic waves. 
After the warm-up period, the speed at which the ego vehicle is driven at is determined by the driver policy acting upon the advised action.
The action is provided to SUMO and is then propagated in simulation.

Intuitively, personalized PC policies could be trained from scratch. 
However, the training time for these PC policies makes this approach inefficient. 
We aim to improve performance by using the already trained PC policies without additional overhead in terms of training time.
We show that our PeRP achieves improvements while being trained for significantly fewer steps than the PC policy, thus imposing minimal overhead on training.

Finally, we note that PeRP is also a PC policy as it assumes that actions will be held for $\delta$ timesteps to account for driver reaction and adjustment periods.  
We assume that the driver trait does not change during the hold-length as it is unlikely for human traits to change within such a short duration. 



\begin{algorithm}[tb!]
\caption{The PeRP algorithm}
\label{alg:perp}
\DontPrintSemicolon
\KwIn{Piecewise constant policy: $\pi^{PCP}$, Driver policy: $\pi^{driver}$, Driver Trait Inference model: $f_{\text{trait}}$, Learning rate: $\lambda$, Hold Length: $\delta$, Warm-up time: $W$, Horizon: $H$, Observation length: $T$, Maximum training iterations: $n_{max}$}
Initialize $\pi^{PeRP}_\theta$\;
$iter \gets 0$\;
\While{$iter \le n_{max}$ and not converged}{
    $s_t \gets$ Reset environment\;
    \For{$t \in \{1, ..., W\}$}{
        Sample $a_t \sim IDM(s_t)$\;
        Execute action $a_t$ in environment\;
        Observe $s_{t + 1}$\;
    }
    \While{$t \le W+H$}{
        Sample $a^{PCP}_t \sim \pi^{PCP}(s_t)$\;
        $z_t \gets f_{\text{trait}}(\{s_{t-T}, ..., s_{t}\})$\;
        Sample $a^{PeRP}_t \sim \pi^{PeRP}_\theta(s_t, a^{PCP}_t, z_t)$\;
        $a^{advised}_t \gets a^{PCP}_t + a^{PeRP}_t$\;
        \For{$\Tilde{t} \in \{1, ..., \delta\}$}{
            Sample $a_{t + \Tilde{t}} \sim \pi^{driver}(s_{t + \Tilde{t}}, a^{advised}_{t + \Tilde{t}})$\;
            Execute action $a_{t + \Tilde{t}}$ in environment\;
            Observe $s_{{t + \Tilde{t}} + 1}, r_{t + \Tilde{t}}$\;
        }
        $t \gets t + \delta$\;
    }
    Estimate loss
    $L_\theta(\theta)$ using $\{s_W, ..., s_{W+H}\}$, $\{a_W, ..., a_{W+H}\}$, and $\{r_W, ..., r_{W+H}\}$\;
    $\theta \gets \theta - \lambda \nabla_\theta L_\theta(\theta)$\;
    $iter \gets iter + 1$
}
\end{algorithm}


%% file: sections/04_experiments.tex
\section{Experiments}
\label{sec:experiments}


\subsection{Environment Setup}

All experiments were carried out in simulation\footnotemark[1] using the Flow framework for RL based on SUMO~\cite{wu2022flow, sumo}.
Every timestep in the simulation was equivalent to 0.1 seconds.
The actions implemented by the agent are provided to SUMO and are internally propagated to take effect in the next step. 
We set the $min_{gap}$ parameter for the ego vehicle to be 0. 
Thus, the policy is required to learn to avoid collisions with the leader by controlling the speed of the agent.
If a collision occurs during a rollout, the episode is ended with 0 reward.

Our experiments involve vehicles driving on the canonical single-lane circular track as is standard in many related works~\cite{pcp, stern2018}.
Our track has a circumference of 640m with $N=40$ vehicles on the road.
The ego vehicle is controlled by the driver model detailed in Section \ref{sec:method-driver-policy} following the instructions of the policy.
The other 39 "human" drivers on the track are controlled by IDM~\cite{idm}. 
Each vehicle on the track is 5m long.
We assume that every vehicle always has a leader.
The maximum speed, $A_{max}$, was set to 35m/s based on empirical simulation results for all policies as it is unlikely for vehicles to reach this speed given the initial conditions.

Even though the PeRP is designed to handle changing driver traits, the traits were kept constant for a rollout to prevent drastic changes leading to unrealistic simulation \eg: the trait shifting from $\text{trait}_\mu = 5$ to $\text{trait}_\mu = -5$ between consecutive action extension periods. 

We carried out analysis and experiments for hold-lengths: $\delta \in \{10, 20, 50, 100\}$. 
We choose these particular values to provide a wide range of possible extensions to test our system at the extremes.

\subsection{Piecewise Constant Policies}
Policies were trained for each $\delta$ with $\alpha=18$ actions and $A_{max}=35$~m/s in the action space $A^{PCP}$.
Each policy was trained as a multi-layer perceptron, with hidden layers of shape $(64, 64)$ for 1000 iterations using TRPO~\cite{schulman2015trpo}.
Each iteration had a warm-up period of 1000 steps and a horizon of 2000 steps.
A $\gamma = 0.99$ and a learning rate of $0.0001$ resulted in the best performing policies.
Only the best performing policies for each $\delta$ were used as bases for DTI and PeRP.
All other parameters were kept consistent with those presented by Sridhar \etal~\cite{pcp}.
All PC policies were trained using an Intel Xeon Platinum 8260 processor and 4 CPUs provided by the MIT SuperCloud~\cite{reuther2018interactive}.

\subsection{Driver Trait Inference}
\textbf{Dataset:} We train the DTI model on a dataset collected in simulation using the driver policy model introduced in Section \ref{sec:method-driver-policy} following advice from the PC policies trained above. 
All states after the warm-up period of 600 steps are collected for 50 iterations for a horizon of 1000 steps for each PCP.
In total, we collected 36,750 driving trajectories of length $T=20$ with equal distribution amongst the five driving traits.
We perform an 80-20 train-evaluation dataset split for the collected trajectories.

\textbf{Training:} 
We train our VAE model with a latent space size of $l=2$, with both encoder and decoder networks as single layer LSTMs with hidden and cell sizes of 32. 
Our best performing model was obtained by training for 100 epochs or until convergence with learning rate of 0.0001, batch size of 16, $\beta_{\text{recon}} = 1$, and $\beta_{\text{KL}} = 0.0001$.
All models were trained on the HAL Cluster using a single Nvidia V100 GPU~\cite{hal}.

\begin{table*}[b!]
    \centering
    \caption{Results comparing the performance of our baselines with PeRP}
    \resizebox{\textwidth}{!}{\begin{tabular}{l  c c  c  c c  c  c c  c  c c}
        \toprule
         Policy & 
         \multicolumn{2}{c}{$\delta=10~(1\text{s})$} && 
         \multicolumn{2}{c}{$\delta=20~(2\text{s})$} && 
         \multicolumn{2}{c}{$\delta=50~(5\text{s})$} &&
         \multicolumn{2}{c}{$\delta=100~(10\text{s})$} \\
         \cmidrule{2-3} \cmidrule{5-6} \cmidrule{8-9} \cmidrule{11-12} 
         & Avg. Speed$(\uparrow)$& Avg. Std$(\downarrow)$&
         & Avg. Speed$(\uparrow)$& Avg. Std$(\downarrow)$&
         & Avg. Speed$(\uparrow)$ & Avg. Std$(\downarrow)$&
         & Avg. Speed$(\uparrow)$& Avg. Std$(\downarrow)$\\
         \midrule
         Optimal & 
         8.650 & - &&
         - & - &&
         - & - &&
         -  & -  \\
         OSL & 
         6.764 & 1.768 &&
         - & - &&
         - & - &&
         -  & -  \\
         PCP & 
         7.891 & 1.477 && 
         8.206 & 1.635 && 
         7.527 & \textbf{1.901} && 
         6.449 & 1.847 \\ 
         V-RP &
         8.031 & 1.513 && 
         8.220 & 1.620 && 
         7.638 & 2.023 && 
         6.576 & 1.944 \\ 
         TA-RP   & 
         8.055 & 1.454 && 
         8.222 & 1.627 && 
         7.816 & 1.967 && 
         6.467 & 2.016 \\ 
         \textbf{PeRP$^*$} & 
         \textbf{8.077} & \textbf{1.449} && 
         \textbf{8.235} & \textbf{1.588} && 
         \textbf{7.848} & 1.953 && 
         \textbf{6.704} & \textbf{1.842} \\ 
        \bottomrule
    \end{tabular}}
    \label{tab:quantitative}
\end{table*}

\subsection{Personalized Residual Policies}
\textbf{Training:} The PeRPs were trained and evaluated for each of the PC policies with a warm-up period of 600 steps and a horizon of 4000 steps.
Similar to PC policies, each PeRP was trained as a Multi-layer perceptron, with hidden layers of shape $(64, 64)$.
All policies were trained for 200 iterations using TRPO~\cite{schulman2015trpo}.  
A $\gamma = 0.99$ and a learning rate of $0.0001$ resulted in the best performing policies.
The action space bound, $\epsilon$, for $A^{PeRP}$ was set to 6m/s as values between $[-6, 6]$ capture 87\% of the the driver policies offset. 
Lastly, PeRP uses an observation window of $T=20$ for the trait inference.
We note that while the DTI model and PeRP could be trained simultaneously, they were trained separately to avoid compounding errors. 

\textbf{Baselines:}
We evaluate our PeRPs against four baselines:
\textbf{(1) Optimal Speed Limit (OSL)}: The policy that outputs a constant action calculated based on the network parameters verified to maximize the average speed of all vehicles. For our network, this action was a speed of 8.65 m/s. This policy is equal to one that is trained for $\delta=\infty$ and is analogous to a policy that always advises the speed limit;
\textbf{(2) Piecewise Constant Policy}: The base policy for PeRP and the other RL based baselines; 
\textbf{(3) Vanilla Residual Policy (V-RP)}: A simple residual policy that aims at offsetting the base PC policy without any trait knowledge;
\textbf{(4) Trait Aware Residual Policy (TA-RP)}: A residual policy conditioned on the ground truth driver trait. We expect this model to have the best performance due to its ground truth input.

\textbf{Metrics:}
We evaluate the PeRPs and their baseline variants on two metrics: 
\textbf{(1) Average Speed:} The average speed of all the vehicles on the road. A standard metric where a high average speed indicates less congestion;
\textbf{(2) Average Standard Deviation of Speed:} The average of the standard deviation in speed over all iterations. A lower value indicates less congestion. 

%% file: sections/05_discussion.tex
\section{Results and Discussion}
\label{sec:results}

\subsection{Driver Trait Inference}
\begin{figure}
    \centering
    \includegraphics[width=\columnwidth]{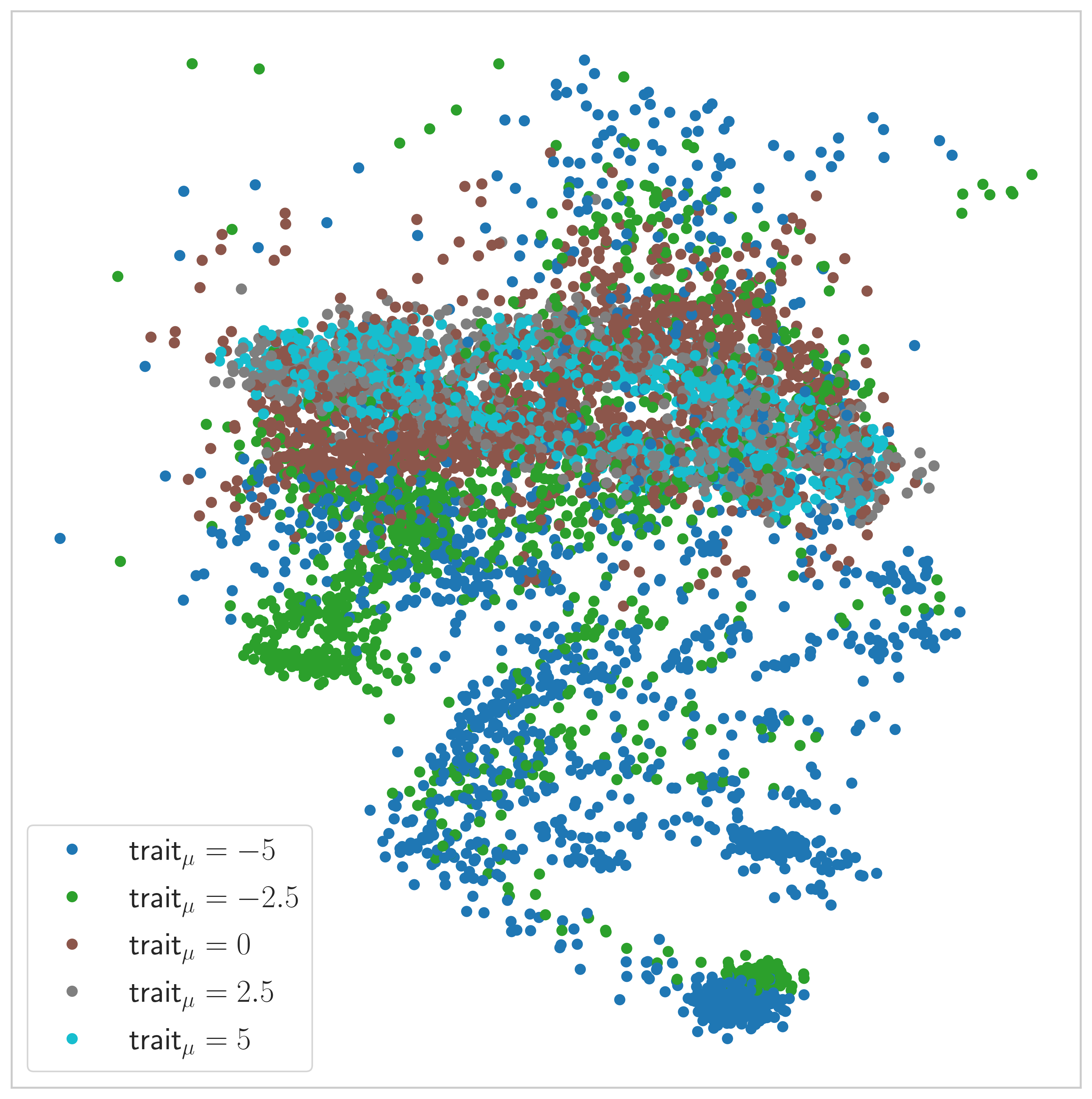}
    \caption{A visualisation of the latent space of the driver trait inference VAE on the validation dataset. We show the latent points for the different trait means, $\text{trait}_\mu$, in different colors.}
    \label{fig:vae_latent}
\end{figure}

Figure \ref{fig:vae_latent} shows the latent space for the validation dataset.
The VAE model is successful in clustering the trajectories of the different driving styles.
We do not achieve fully separate clusters as is expected with unsupervised models.
However, the clusters for the different traits are noticeably distinct with minor overlap.
We omit an analysis of the reconstructed trajectories as they have little impact on the PeRPs.

\subsection{Personalized Residual Policies}
Table \ref{tab:quantitative} shows our quantitative results for the best performing models evaluated for 100 iterations. 
Note that, the same PC policy was used for each $\delta$ for all the baselines dependent upon a PC policy \ie~the V-RP, TA-RP, and PeRP were trained using the same PC policy that they are compared against for a particular $\delta$.

We observe that PeRP indeed outperforms the baseline models as evidenced by the high average speeds with low average standard deviations in speed as seen in Table \ref{tab:quantitative}.
The residual policies with driver trait inference, TA-RP, do not perform proportionately with PeRP. 
We suspect that this under-performance is due to differences in exploration during training.
Meanwhile, all policies significantly outperform the OSL.
This difference in performance can be attributed to the online nature of the RL policies that adapt guidance rather than advising a constant action like the OSL.
In addition, all policies except OSL successfully avoid collisions. 
Some baselines incurred collisions in 4\% of evaluation scenarios on average. 
However, PeRP and PCP were both careful to have 0 collisions in all evaluation iterations for $\delta=\{20, 50\}$.
Iterations where collisions occur were omitted during calculation of the metrics shown in Table \ref{tab:quantitative}.
We present collisions and emissions results in supplementary material~\footnotemark[1].

Additionally, the performance of the policies is severely dependent on the hold-length chosen.
A small $\delta$ would place too much load on the driver and change the action too frequently for it to be followed by the driver.
Conversely, a large $\delta$ would require holding an action that is no longer the near optimal action for the driving scenario.
Thus, we conjecture that there exists an optimal $\delta^*$ that allows for drivers to adapt to the advised actions while providing the most updated advice.
From Table \ref{tab:quantitative}, it is clear that such an optimal $\delta^*$ exists when looking at each policy's performance individually. 
In our case, $\delta=20$ that corresponds to a hold-length of 2s.
We speculate further that this optimal $\delta^*$ would depend on the network parameters and the number of cars.

As a hold-length of 1s is impractical, the results for that hold-length do not provide substantial inference when compared to larger hold-lengths.
These results are included solely to allow for comparison with previous work that utilizes AVs. 
For more practical $\delta$s (20, 50), PeRP consistently outperforms the baselines.
As expected, a hold-length of 10s is too large and leads to significant detriment in performance.
All policies also perform well in the case of $\delta=1$\footnotemark[1]. 
This $\delta$ corresponds to a completely online scenario where there is no action extension.
The policies' performance for this $\delta$ showed similar albeit slightly worse performance with metrics aligning with those of $\delta=10$.
However, a $\delta$ of 1 is even more impractical than $\delta=10$ as human controlled actions cannot change in $0.1$s.


PeRP achieves an improvement of 22\% over standard speed limit guidance (OSL) and 4\% over guidance provided by PC policies for $\delta=5$s.
Comparisons on the average standard deviation also show similar results with the PeRP policy for $\delta=2$s achieving improvements of 10\% over OSL and 3\% over PC policies.
While these results are promising and significant, we emphasize that they were obtained completely in simulation, without a human-in-the-loop.
Therefore, we encourage the evaluation of these policies with user studies for further analysis.

         
         
         
         

%% file: sections/06_conclusion.tex
\section{Conclusion and Future Work}
\label{sec:conclusion}

In this paper, we present PeRP, a novel Personalized Residual Policies for congestion mitigation. 
We demonstrate the efficacy of our model in a robust simulation study and showcase its ability to adapt to different driving styles.
Particularly, we utilize a novel driver trait inference model to capture nuances in driver behaviors towards following instructions to condition our residual policy.
Our PeRPs show improvements over all baselines, including PC Policies.

This advancement was built on four assumptions which can be relaxed to open the following avenues for future work:
(1) More robust policies should be trained with careful modeling for the driver policy, $\pi^{driver}$, that consider driver reaction times and distractions;  
(2) As Piecewise Constant Policies have empirically been shown to be robust to lane changes, extending PeRP to account for these changes would further increase its applicability;
(3) 
Designing and testing PeRPs where multiple agents receive instructions simultaneously would drastically improve congestion mitigation and would provide a realizable bridge towards a fully AV controlled traffic setting;
(4) As our policies are trained and evaluated completely in simulation, a human-in-the-loop user study would provide further credence to the claims showcased here. 
While naturalistic driving studies would be ideal, driving simulator studies can be performed without considerable effort using the CARLA driving simulator~\cite{carla}.
Particularly, works that have integrated the flow framework with CARLA 
can be used to easily test our model~\cite{hcd-workshop}.
As shown in this paper, future work in advisory algorithms can mitigate congestion by compensating for users' diverse following behavior.



%% file: sections/09_appendix.tex

%% file: root.bbl
\begin{thebibliography}{10}
\providecommand{\url}[1]{#1}
\csname url@rmstyle\endcsname
\providecommand{\newblock}{\relax}
\providecommand{\bibinfo}[2]{#2}
\providecommand\BIBentrySTDinterwordspacing{\spaceskip=0pt\relax}
\providecommand\BIBentryALTinterwordstretchfactor{4}
\providecommand\BIBentryALTinterwordspacing{\spaceskip=\fontdimen2\font plus
\BIBentryALTinterwordstretchfactor\fontdimen3\font minus
  \fontdimen4\font\relax}
\providecommand\BIBforeignlanguage[2]{{%
\expandafter\ifx\csname l@#1\endcsname\relax
\typeout{** WARNING: IEEEtran.bst: No hyphenation pattern has been}%
\typeout{** loaded for the language `#1'. Using the pattern for}%
\typeout{** the default language instead.}%
\else
\language=\csname l@#1\endcsname
\fi
#2}}

\bibitem{epa}
{United States. Environmental Protection Agency. Office of Policy},
  \emph{Inventory of US greenhouse gas emissions and sinks: 1990-2020}.\hskip
  1em plus 0.5em minus 0.4em\relax United States Environment Protection Agency,
  2020.

\bibitem{emissions}
M.~Barth and K.~Boriboonsomsin, ``Real-world carbon dioxide impacts of traffic
  congestion,'' \emph{Transportation Research Record}, vol. 2058, pp. 163--171,
  2008.

\bibitem{goodwin2004economic}
P.~Goodwin, ``The economic costs of road traffic congestion,'' 2004.

\bibitem{french}
L.~Carnis and E.~Blais, ``An assessment of the safety effects of the french
  speed camera program,'' \emph{Accident Analysis \& Prevention}, vol.~51, pp.
  301--309, 2013.

\bibitem{wilmot1999}
C.~G. Wilmot and M.~Khanal, ``Effect of speed limits on speed and safety: A
  review,'' \emph{Transport Reviews}, vol.~19, no.~4, pp. 315--329, 1999.

\bibitem{flow}
C.~Wu, A.~R. Kreidieh, K.~Parvate, E.~Vinitsky, and A.~M. Bayen, ``Flow: A
  modular learning framework for mixed autonomy traffic,'' \emph{IEEE
  Transactions on Robotics}, 2021.

\bibitem{stern2018}
R.~E. Stern, S.~Cui, M.~L. {Delle Monache}, R.~Bhadani, M.~Bunting,
  M.~Churchill, N.~Hamilton, R.~Haulcy, H.~Pohlmann, F.~Wu, B.~Piccoli,
  B.~Seibold, J.~Sprinkle, and D.~B. Work, ``Dissipation of stop-and-go waves
  via control of autonomous vehicles: Field experiments,'' \emph{Transportation
  Research Part C: Emerging Technologies}, vol.~89, pp. 205--221, 2018.

\bibitem{Kalra2014}
N.~Kalra and S.~M. Paddock, \emph{Driving to Safety: How Many Miles of Driving
  Would It Take to Demonstrate Autonomous Vehicle Reliability?}\hskip 1em plus
  0.5em minus 0.4em\relax RAND Corporation, 2016.

\bibitem{dixit2016autonomous}
V.~V. Dixit, S.~Chand, and D.~J. Nair, ``Autonomous vehicles: Disengagements,
  accidents and reaction times,'' \emph{PLOS ONE}, vol.~11, no.~12, pp. 1--14,
  12 2016.

\bibitem{shared_control_survey}
M.~Marcano, S.~Díaz, J.~Pérez, and E.~Irigoyen, ``A review of shared control
  for automated vehicles: Theory and applications,'' \emph{IEEE Transactions on
  Human-Machine Systems}, vol.~50, pp. 475--491, 2020.

\bibitem{pcp}
M.~Sridhar and C.~Wu, ``Piecewise constant policies for human-compatible
  congestion mitigation,'' in \emph{IEEE International Intelligent
  Transportation Systems Conference}, 2021.

\bibitem{emergent}
C.~Wu, A.~Kreidieh, E.~Vinitsky, and A.~M. Bayen, ``Emergent behaviors in
  mixed-autonomy traffic,'' in \emph{Conference on Robot Learning}, 2017.

\bibitem{kreidieh_dissipating_2018}
A.~R. Kreidieh, C.~Wu, and A.~M. Bayen, ``Dissipating stop-and-go waves in
  closed and open networks via deep reinforcement learning,'' in \emph{IEEE
  {International} {Conference} on {Intelligent} {Transportation} {Systems}},
  Nov 2018, pp. 1475--1480.

\bibitem{vinitsky_lagrangian_2018}
E.~Vinitsky, K.~Parvate, A.~Kreidieh, C.~Wu, and A.~Bayen, ``Lagrangian
  {Control} through {Deep}-{RL}: {Applications} to {Bottleneck}
  {Decongestion},'' in \emph{IEEE {International} {Conference} on {Intelligent}
  {Transportation} {Systems}}, Nov 2018, pp. 759--765.

\bibitem{yan_reinforcement_2021}
Z.~Yan and C.~Wu, ``Reinforcement {Learning} for {Mixed} {Autonomy}
  {Intersections},'' in \emph{{IEEE} {International} {Intelligent}
  {Transportation} {Systems} {Conference}}, Sep 2021, pp. 2089--2094.

\bibitem{jayawardana2022ecolearning}
V.~Jayawardana and C.~Wu, ``Reinforcement learning for eco-lagrangian control
  at intersections,'' in \emph{European Control Conference}, 2022.

\bibitem{jayawardana_mixed_2021}
V.~Jayawardana, A.~Landler, and C.~Wu, ``Mixed {Autonomous} {Supervision} in
  {Traffic} {Signal} {Control},'' in \emph{{IEEE} {International} {Intelligent}
  {Transportation} {Systems} {Conference}}, Sep 2021, pp. 1767--1773.

\bibitem{wegener_automated_2021}
M.~Wegener, L.~Koch, M.~Eisenbarth, and J.~Andert, ``Automated eco-driving in
  urban scenarios using deep reinforcement learning,'' \emph{Transportation
  Research Part C: Emerging Technologies}, vol. 126, p. 102967, 2021.

\bibitem{metelli_control_2020}
A.~M. Metelli, F.~Mazzolini, L.~Bisi, L.~Sabbioni, and M.~Restelli, ``Control
  {Frequency} {Adaptation} via {Action} {Persistence} in {Batch}
  {Reinforcement} {Learning},'' in \emph{Proceedings of the 37th
  {International} {Conference} on {Machine} {Learning}}, Nov 2020, pp.
  6862--6873.

\bibitem{li_integrated_2023}
\BIBentryALTinterwordspacing
S.~Li, R.~Dong, and C.~Wu, ``Integrated {Analysis} of {Human}-compatible
  {Control} for {Traffic} {Flow} {Stability},'' Jan 2023. [Online]. Available:
  \url{http://arxiv.org/abs/2301.04043}
\BIBentrySTDinterwordspacing

\bibitem{10.1145/1143844.1143845}
P.~Abbeel, M.~Quigley, and A.~Y. Ng, ``Using inaccurate models in reinforcement
  learning,'' in \emph{Proceedings of the 23rd International Conference on
  Machine Learning}, 2006.

\bibitem{pmlr-v168-jiang22a}
Y.~Jiang, J.~Sun, and C.~K. Liu, ``Data-augmented contact model for rigid body
  simulation,'' in \emph{Proceedings of The 4th Annual Learning for Dynamics
  and Control Conference}, ser. Proceedings of Machine Learning Research, vol.
  168, 23--24 Jun 2022, pp. 378--390.

\bibitem{10.1109/TRO.2020.2988642}
A.~Zeng, S.~Song, J.~Lee, A.~Rodriguez, and T.~Funkhouser, ``Tossingbot:
  Learning to throw arbitrary objects with residual physics,'' \emph{Trans.
  Rob.}, vol.~36, no.~4, p. 1307–1319, Aug 2020.

\bibitem{Ajay2018AugmentingPS}
A.~Ajay, J.~Wu, N.~Fazeli, M.~Bauz{\'a}, L.~P. Kaelbling, J.~B. Tenenbaum, and
  A.~Rodriguez, ``Augmenting physical simulators with stochastic neural
  networks: Case study of planar pushing and bouncing,'' \emph{IEEE/RSJ
  International Conference on Intelligent Robots and Systems}, pp. 3066--3073,
  2018.

\bibitem{schaff2020}
C.~Schaff and M.~R. Walter, ``Residual policy learning for shared autonomy,''
  \emph{Proceedings of Robotics: Science and Systems}, 2020.

\bibitem{silver2018}
T.~Silver, K.~R. Allen, J.~Tenenbaum, and L.~P. Kaelbling, ``Residual policy
  learning,'' \emph{CoRR}, vol. abs/1812.06298, 2018.

\bibitem{johannik2019}
T.~Johannink, S.~Bahl, A.~Nair, J.~Luo, A.~Kumar, M.~Loskyll, J.~A. Ojea,
  E.~Solowjow, and S.~Levine, ``Residual reinforcement learning for robot
  control,'' in \emph{2019 International Conference on Robotics and
  Automation}, 2019.

\bibitem{pmlr-v205-rana23a}
K.~Rana, M.~Xu, B.~Tidd, M.~Milford, and N.~Suenderhauf, ``Residual skill
  policies: Learning an adaptable skill-based action space for reinforcement
  learning for robotics,'' in \emph{Proceedings of The 6th Conference on Robot
  Learning}, ser. Proceedings of Machine Learning Research, vol. 205, 14--18
  Dec 2023, pp. 2095--2104.

\bibitem{9834085}
R.~Zhang, J.~Hou, G.~Chen, Z.~Li, J.~Chen, and A.~Knoll, ``Residual policy
  learning facilitates efficient model-free autonomous racing,'' \emph{IEEE
  Robotics and Automation Letters}, vol.~7, pp. 11\,625--11\,632, 2022.

\bibitem{brown2020taxonomy}
K.~Brown, K.~Driggs-Campbell, and M.~J. Kochenderfer, ``A taxonomy and review
  of algorithms for modeling and predicting human driver behavior,''
  \emph{arXiv preprint arXiv:2006.08832}, 2020.

\bibitem{Deo2018intent}
N.~Deo, A.~Rangesh, and M.~M. Trivedi, ``How would surround vehicles move? {A}
  unified framework for maneuver classification and motion prediction,''
  \emph{{IEEE} Trans. Intell. Veh.}, vol.~3, no.~2, pp. 129--140, 2018.

\bibitem{Gillmeier2019intent}
K.~Gillmeier, F.~Diederichs, and D.~Spath, ``Prediction of ego vehicle
  trajectories based on driver intention and environmental context,'' in
  \emph{IEEE Intelligent Vehicles Symposium}, 2019, pp. 963--968.

\bibitem{huang2020long}
Z.~Huang, A.~Hasan, K.~Shin, R.~Li, and K.~Driggs-Campbell, ``Long-term
  pedestrian trajectory prediction using mutable intention filter and warp
  lstm,'' \emph{IEEE Robotics and Automation Letters}, vol.~6, no.~2, pp.
  542--549, 2020.

\bibitem{Liu2019motion}
J.~Liu, Y.~Luo, H.~Xiong, T.~Wang, H.~Huang, and Z.~Zhong, ``An integrated
  approach to probabilistic vehicle trajectory prediction via driver
  characteristic and intention estimation,'' in \emph{IEEE Intelligent
  Transportation Systems Conference}, 2019, pp. 3526--3532.

\bibitem{morton2017}
J.~Morton and M.~J. Kochenderfer, ``Simultaneous policy learning and latent
  state inference for imitating driver behavior,'' in \emph{IEEE International
  Conference on Intelligent Transportation Systems}, 2017.

\bibitem{liu2021learning}
S.~Liu, P.~Chang, H.~Chen, N.~Chakraborty, and K.~Driggs-Campbell, ``Learning
  to navigate intersections with unsupervised driver trait inference,'' in
  \emph{IEEE International Conference on Robotics and Automation}, 2022.

\bibitem{ma2020reinforcement}
X.~Ma, J.~Li, M.~J. Kochenderfer, D.~Isele, and K.~Fujimura, ``Reinforcement
  learning for autonomous driving with latent state inference and
  spatial-temporal relationships,'' in \emph{IEEE International Conference on
  Robotics and Automation}, 2021.

\bibitem{lili}
A.~Xie, D.~Losey, R.~Tolsma, C.~Finn, and D.~Sadigh, ``Learning latent
  representations to influence multi-agent interaction,'' in \emph{Proceedings
  of the 2020 Conference on Robot Learning}, ser. Proceedings of Machine
  Learning Research, vol. 155, 16--18 Nov 2021, pp. 575--588.

\bibitem{wang2021influencing}
W.~Z. Wang and A.~Shih, ``Influencing towards stable multi-agent
  interactions,'' in \emph{Conference on Robot Learning}, 2021.

\bibitem{losey_rili}
S.~Parekh, S.~Habibian, and D.~P. Losey, ``Rili: Robustly influencing latent
  intent,'' 2022.

\bibitem{vae}
D.~Kingma and M.~Welling, ``Auto-encoding variational bayes.'' in
  \emph{Conference on Learning Representations}, 2014.

\bibitem{idm}
M.~Treiber, A.~Hennecke, and D.~Helbing, ``Congested traffic states in
  empirical observations and microscopic simulations,'' \emph{Physical review
  E}, vol.~62, no.~2, p. 1805, 2000.

\bibitem{Droździel2020}
P.~Droździel, S.~Tarkowski, I.~Rybicka, and R.~Wrona, ``Drivers’ reaction
  time research in the conditions in the real traffic,'' \emph{Open
  Engineering}, vol.~10, no.~1, pp. 35--47, 2020.

\bibitem{Jurecki2017}
R.~S. Jurecki, T.~L. Stańczyk, and M.~J. Jaśkiewicz, ``Driver’s reaction
  time in a simulated, complex road incident,'' \emph{Transport}, vol.~32,
  no.~1, pp. 44--54, 2017.

\bibitem{hcd-workshop}
A.~{Hasan}, N.~{Chakraborty}, C.~{Wu}, and K.~{Driggs-Campbell}, ``Towards
  co-operative congestion mitigation,'' in \emph{Proceedings of the `Shared
  Autonomy in Physical Human-Robot Interaction: Adaptability and Trust
  Workshop` at the IEEE International Conference on Robotics and Automation},
  2022.

\bibitem{mannering2009}
F.~Mannering, ``An empirical analysis of driver perceptions of the relationship
  between speed limits and safety,'' \emph{Transportation Research Part F:
  Traffic Psychology and Behaviour}, vol.~12, pp. 99--106, 2009.

\bibitem{hasan2022metapath}
A.~Hasan, P.~Sriram, and K.~Driggs-Campbell, ``Meta-path analysis on
  spatio-temporal graphs for pedestrian trajectory prediction,'' in \emph{IEEE
  International Conference on Robotics and Automation}, 2022.

\bibitem{wu2022flow}
C.~Wu, A.~R. Kreidieh, K.~Parvate, E.~Vinitsky, and A.~M. Bayen, ``Flow: A
  modular learning framework for mixed autonomy traffic,'' \emph{IEEE
  Transactions on Robotics}, vol.~38, no.~2, pp. 1270--1286, 2022.

\bibitem{sumo}
P.~A. Lopez, M.~Behrisch, L.~Bieker-Walz, J.~Erdmann, Y.-P. Fl{\"o}tter{\"o}d,
  R.~Hilbrich, L.~L{\"u}cken, J.~Rummel, P.~Wagner, and E.~Wie{\ss}ner,
  ``Microscopic traffic simulation using sumo,'' in \emph{The 21st IEEE
  International Conference on Intelligent Transportation Systems}, 2018.

\bibitem{schulman2015trpo}
J.~Schulman, S.~Levine, P.~Abbeel, M.~Jordan, and P.~Moritz, ``Trust region
  policy optimization,'' in \emph{Proceedings of the 32nd International
  Conference on Machine Learning}, ser. Proceedings of Machine Learning
  Research, vol.~37, 07--09 Jul 2015, pp. 1889--1897.

\bibitem{reuther2018interactive}
A.~Reuther, J.~Kepner, C.~Byun, S.~Samsi, W.~Arcand, D.~Bestor, B.~Bergeron,
  V.~Gadepally, M.~Houle, M.~Hubbell, M.~Jones, A.~Klein, L.~Milechin,
  J.~Mullen, A.~Prout, A.~Rosa, C.~Yee, and P.~Michaleas, ``Interactive
  supercomputing on 40,000 cores for machine learning and data analysis,'' in
  \emph{2018 IEEE High Performance extreme Computing Conference}, 2018, pp.
  1--6.

\bibitem{hal}
V.~Kindratenko, D.~Mu, Y.~Zhan, J.~Maloney, S.~H. Hashemi, B.~Rabe, K.~Xu,
  R.~Campbell, J.~Peng, and W.~Gropp, ``Hal: Computer system for scalable deep
  learning,'' in \emph{Practice and Experience in Advanced Research Computing},
  ser. Pearc '20, 2020, p. 41–48.

\bibitem{carla}
A.~Dosovitskiy, G.~Ros, F.~Codevilla, A.~Lopez, and V.~Koltun, ``{CARLA}: {An}
  open urban driving simulator,'' in \emph{Proceedings of the 1st Annual
  Conference on Robot Learning}, 2017, pp. 1--16.

\end{thebibliography}
